# Surface abnormality detection in medical and inspection systems using energy variations in co-occurrence matrixes

**Nandara K. Krishnand[1], Akshakhi Kumar Pritoonka[1], Faeze Kiani[2,*]**
[1]Department of textural science, Online Computer Vision Research center, India
[2]Department of electronic science, Online Computer Vision Research center, Iran
[*]ocvrgroup000@gmail.com

**Abstract:**

Detection of surface defects is one of the most important issues in the field of image processing and machine vision. In this article, a method for detecting surface defects based on energy changes in co-occurrence matrices is presented. The presented method consists of two stages of training and testing. In the training phase, the co-occurrence matrix operator is first applied on healthy images and then the amount of output energy is calculated. In the following, according to the changes in the amount of energy, a suitable feature vector is defined, and with the help of it, a suitable threshold for the health of the images is obtained. Then, in the test phase, with the help of the calculated quorum, the defective parts are distinguished from the healthy ones. In the results section, the mentioned method has been applied on stone and ceramic images and its detection accuracy has been calculated and compared with some previous methods. Among the advantages of the presented method, we can mention high accuracy, low calculations and compatibility with all types of levels due to the use of the training stage. The proposed approach can be used in medical applications to detect abnormalities such as diseases. So, the performance is evaluated on 2d-hela dataset to classify cell phenotypes. The proposed approach provides about 89.56 percent accuracy on 2d-hela.

## 1. Introduction

The visible surface of any object is called its surface. Therefore, any defect that changes the appearance of the surface (surface) and creates non-normality is called surface defects. Detection of surface defects is very important in various fields such as factory production, quality control and medicine. Therefore, researchers are always trying to design intelligent systems to detect surface defects that can perform this operation in less time and with higher accuracy. The most famous intelligent defect detection systems, sometimes called automatic inspection systems, are based on image processing and machine vision techniques. For example, in [1], a method for detecting wood surface defects is presented. Gash and his colleagues in [2] proposed a fully automatic method for detecting fabric defects, and similarly, algorithms for leather [3], agricultural products [4] and metal plates [5] have been mentioned so far.

Most of the methods that have been presented so far are designed only for a specific product or application (ceramic, paper, leather, etc.) Therefore, in this article, a method for detecting defects is presented, which can be used for most applications without loss of accuracy by using one training step.

In most of the proposed techniques, they try to analyze the texture of the images first and define appropriate feature vectors to introduce that texture. Finally, by extracting feature vectors from images and using classifiers, they do the diagnosis. Therefore, as seen, the most important step in defect detection is texture analysis and definition of a suitable feature vector that can be used to distinguish different types of textures from each other.

Tissue analysis techniques [6] can be divided into 4 major groups. Statistical techniques [7], structural [8], model-based [9] and text-based filter [10]. Co-occurrence matrices are one of the statistical techniques. In this article, the desired images are first analyzed by co-occurrence matrices in different directions, and then the energy level of the output images is calculated in different directions. In the following, by comparing the amount of energy in different states, a suitable feature vector that is a good representative of the texture of that image is defined. In the following, according to the defining feature vector, a two-step method for detecting defects is presented. The training phase consists of windowing perfectly intact images and extracting feature vectors from them. Also, at this stage, by obtaining the average of the extracted vectors and calculating the distance of each of the windows with the average vector, the quorum of the windows is obtained. Finally, in the test phase, by windowing the images and comparing with the quorum, surface defects are revealed and detected. In the results section, to determine the "diagnosis accuracy" of the presented method, images of surface defects on agricultural products, stone and ceramics were collected and the method was applied to them. High detection accuracy, lack of dependence on the application context, and low calculations are among the advantages of the method, which are discussed in the results section.The proposed approach is a general algorithm which is learned the defect. So, it can be used in medical applications such as bacteria abnormality detection [15], DNA structure [16], etc.

## 2. Co-occurrence matrix

Co-occurrence matrices were first proposed by Harlick in [11]. The co-occurrence matrix is a quadratic operator that examines the spatial relationship of each pixel of the image with its neighbors. The co-occurrence matrix operator calculates that in the input image, the brightness intensity "j" has been seen several times after the brightness intensity "i" according to the defined spatial relation. This issue is shown in figure (1). As seen in Figure (1-A), the input image is 3 levels and the spatial relationship is defined as the first pixel on the right. Then, the co-occurrence matrix of the input image is calculated in Figure (1-b). For example, the number 3 in the second row and the first column shows that 3 times the brightness of zero has occurred in the pixel to the right of the brightness of one.

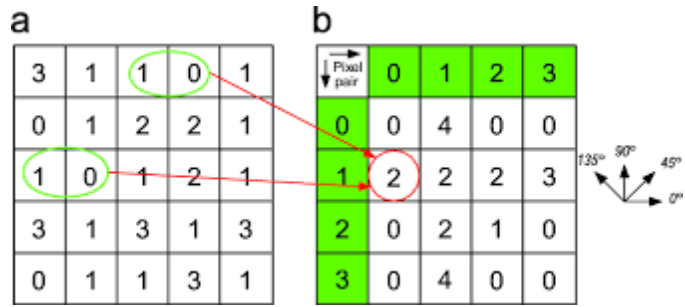

Figure 1. GLCM process numerical example
a) original image b)produced GLCM

According to the above explanations, the co-occurrence matrix of each image can be calculated according to different spatial relations. In this regard, generally 8 types of spatial relationships are considered for calculating the co-occurrence matrix, which are also called 8 directions according to their degree of rotation with respect to the original pixel. This issue is shown in equation 1.

$$R=[i,j+1] \ 0° \ R=[i-1,j+1] \ 45° \ R=[i-1,j] \ 90° \ R=[i-1,j-1] \ 135°$$
$$R=[i,j-1] \ 180° \ R=[i+1,j-1] \ 225° \ R=[i+1,j] \ 270° \ R=[i+1,j=1] \ 315° \quad (1)$$

After calculating the co-occurrence matrix, the obtained matrix can be shown in the form of an output image. In figure (2), the image of the co-occurrence matrix calculated for an image of orange travertine stone texture is shown.

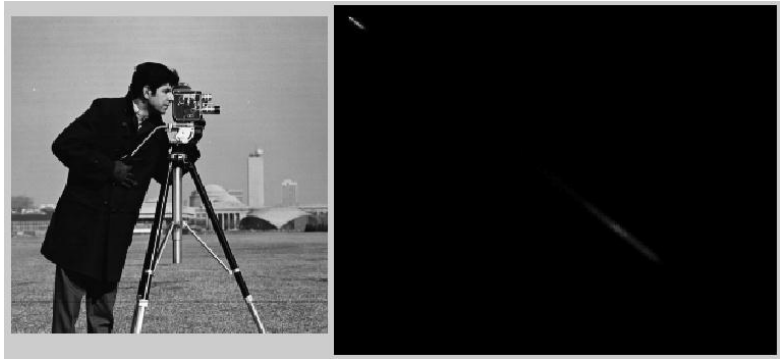

Figure 2. Output example of GLCM in image format

(a) Original image (b) GLCM output in degree zero

## 3. Feature extraction phase

One of the important statistical features proposed in [12] for image analysis is the amount of image energy. The amount of energy of the image can provide useful information about how and how the brightness intensities of the images are scattered. In this regard, to calculate the energy, first the histogram of the image is calculated and normalized. Then, with the help of equation (2), the energy of the image can be calculated. In equation (2), P(g) shows the probability of hitting the light intensity with the gray level (g), which is the height of house g in the normalized histogram.

$$Energy = \sum_{g=0}^{L-1}[p(g)]^2 \quad (2)$$

Now, in order to define the special feature vector of this problem, the following algorithm is presented:

A) Calculation of the co-occurrence matrix for the input image in zero, 45, 90, 135 directions

b) Calculating the amount of energy of the 4 images of the obtained co-occurrence matrix

c) Defining the 6-dimensional feature vector, each dimension of which is the amount of energy changes of the matrices with respect to each other. Equation (3) shows the defining feature vector

$$F=<(E(0°)-E(45°)),(E(0°)-E(90°)),(E(0°)-E(135°)),(E(45°)-E(90°)),(E(45°)-E(135°)),(E(90°)-E(135°))> \quad (3)$$

In equation (3), E(i) indicates the amount of energy calculated from the image of the co-occurrence matrix obtained in the i-degree direction. It is necessary to explain that the amount of energy of the co-occurrence matrices in the directions zero, 45, 90 and 135 is equal to the energy of the co-occurrence matrices in the directions 180, 225, 270 and 315, respectively. Therefore, only calculating the co-occurrence matrix in 4 directions is sufficient. This point is one of the advantages of the proposed method to reduce computational and time complexity.

## 4. Abnormality detection phase

The main goal of this article is to provide a reliable method for detecting and revealing surface defects. In this regard, a method for detecting surface defects based on energy changes is described in this section. The presented method consists of two stages of training and testing. The purpose of the training phase is to identify and introduce healthy parts to the system. In this regard, first, an image of the examined surface is prepared that is free from defects confirmed by experts. Then the desired image is divided into windows with equal dimensions of W×W. Next, for each of the windows, according to the method presented in section 3, the feature vector is extracted. Now the average vector is calculated according to equation (4). In equation (4), the value of each dimension is equal to the average value of that dimension in all healthy windows.

$$F_{average} = <\sum_{i=1}^{k} F_{1i}, \sum_{i=1}^{k} F_{2i}, \sum_{i=1}^{k} F_{3i}, \sum_{i=1}^{k} F_{4i}, \sum_{i=1}^{k} F_{5i}, \sum_{i=1}^{k} F_{6i}> \quad (4)$$

In the above equation, i represents the i-th window and k is the total number of windows.

Finally, with the help of equation (5), it is possible to obtain a suitable and reliable threshold limit for the healthy level of the healthy window. In equation (5), first the Sorensen distance [13] between all the vectors of the windows is calculated with the mean vector, and then the maximum distance calculated is considered as the threshold limit of the surface being healthy.

$$Threshold = Max\{\frac{\sum_{d=1}^{6}|F_{i_d} - F_{average_d}|}{\sum_{d=1}^{6}|F_{i_d} + F_{average_d}|} \quad i = 1,2,....,k\} \quad (5)$$

With the help of the healthy threshold limit, it is possible to separate the defective parts from the healthy parts in the test phase. In this regard, first the test image is divided into windows with the same W×W dimensions. Then the feature vector is calculated for each of the windows and the distance of each is measured with the average vector (average vector in the training phase) (Equation 5). It is clear that if any of the distances obtained is greater than the healthy threshold, the mentioned window contains a defect and is introduced as a defective window. Threshold can be optimized in different cases such as cancer diagnosis based on abnormalities in cell [17-18], immune human parameter detection [19] inspection systems [20].

## 5. Experimental results

The main goal of this paper was to provide a reliable method for detecting surface defects. In this regard, to check the quality of the presented method, images of surface defects in agricultural products (potatoes), building stones (cream travertine, ax stone) and ceramics were taken with the help of a 16-megapixel camera. Also, some defect-free images were prepared for the training phase.

One of the main criteria used to check the quality of defect detection methods[14] is the detection accuracy criterion. Therefore, after photographing, the mentioned method was applied to the images and the detection accuracy was calculated in each of the images with the help of equation (6). The average accuracy of detection in all images according to application cases is shown in Table (1). Also, two other defect detection methods (Harlick's features, histogram division) were also applied to the images for comparison and the results are entered in Table (1). As can be seen, in most cases, the accuracy of fault detection has improved. It is worth noting that standard deviation was calculated by paired t test method.

$$DR = 100 \times \left(\frac{N_c + N_d}{N_{total}}\right) \quad (6)$$

In equation (6), Nc means the number of truly healthy windows that are also recognized as healthy windows, and Nd represents the number of defective windows that are also introduced and recognized as defective windows. Detection rate is a benchmark criteria which is used most of classification problems in different scopes [21-25].

| Approach<br>Product | Proposed approach | Haralick features | Quantized histogram features |
|---|---|---|---|
| Stone | 92.45 | 81.60 | 86.22 |
| Agricultural products | 95.72 | 88.24 | 96.33 |
| Ceramics | 91.33 | 82.92 | 87.81 |
| Cell phenotypes | 89.56 | 78.64 | 79.35 |

Some examples of the output defect detected pattern is shown in the figure below.

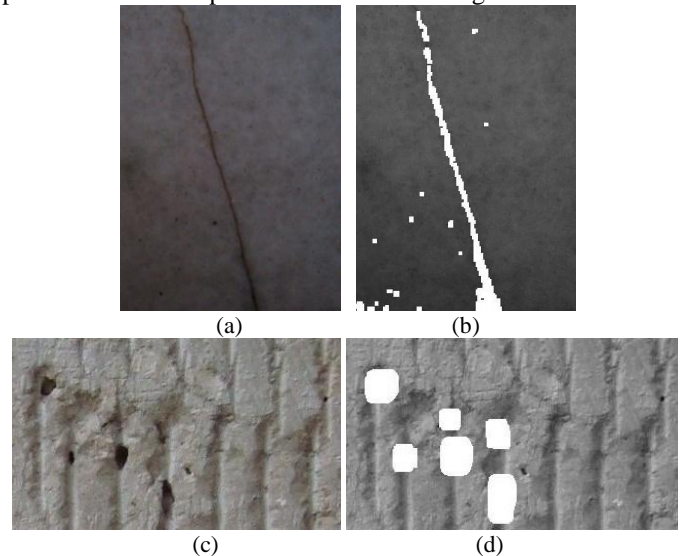

(a)　　　　　(b)

(c)　　　　　(d)

Figure 3. some examples of output defected pattern
(a) original image (b) detected defect pattern output
(c) original image (d) detected defect pattern output

## 6. Discussion and conclusion

The main goal of this paper was to provide a reliable method for detecting surface defects. In this regard, co-occurrence matrix operator was first used for image analysis. Then, with the help of the statistical feature of energy, a suitable feature vector was presented to introduce the texture of the image. In the following, a two-step method for detecting surface defects based on the proposed vector was presented. In the results section, the detection accuracy of the method was checked and compared with some previous methods. The obtained results showed that the presented method has a high ability to detect all types of defects. Among other advantages of the method, we can mention low calculations and insensitivity to noise due to windowing and considering the relationship between pixels. Due to the use of one training step, the presented method can be used in many other fields of application where the main problem is the classification of two classes. Also, the feature vector presented for the first time in this article can be used as a representative vector in many image processing problems such as pattern recognition, object tracking, etc. It is worth noting that the accuracy of the method drops slightly in the face of very small defects, which, of course, can be solved by changing the size of the windows in the windowing stage.